\documentclass[]{fairmeta} % For LaTeX2e

\usepackage{amsmath,amsfonts,bm}

\def\eqref#1{equation~\ref{#1}}
\def\1{\bm{1}}

\DeclareMathAlphabet{\mathsfit}{\encodingdefault}{\sfdefault}{m}{sl}
\SetMathAlphabet{\mathsfit}{bold}{\encodingdefault}{\sfdefault}{bx}{n}

\usepackage[utf8]{inputenc}  % allow utf-8 input
\usepackage[T1]{fontenc}     % use 8-bit T1 fonts
\usepackage{hyperref}        % hyperlinks
\usepackage{url}             % simple URL typesetting
\usepackage{booktabs}        % professional-quality tables
\usepackage{amsfonts}        % blackboard math symbols
\usepackage{nicefrac}        % compact symbols for 1/2, etc.
\usepackage{microtype}       % microtypography
\usepackage{graphicx}
\usepackage{multicol}
\usepackage{multirow}
\usepackage{xspace}
\usepackage{wrapfig}
\usepackage{enumitem}
\usepackage[dvipsnames,table]{xcolor}
\usepackage{colortbl}
\usepackage{caption}
\usepackage{lmodern}
\usepackage{algorithm}
\usepackage{algorithmic}
\usepackage{subcaption}

\usepackage{pifont}
\usepackage{tabularx}
\newcolumntype{R}{>{\raggedleft\arraybackslash}X}
\definecolor{lightblue}{RGB}{220,235,250}
\definecolor{lightgray}{gray}{0.91}
\definecolor{reallab-red}{RGB}{248,191,199}
\definecolor{reallab-blue}{RGB}{59,161,255}
\definecolor{reallab-green}{RGB}{157,207,127}
\definecolor{reallab-yellow}{RGB}{249,180,13}
\usepackage[most,skins,theorems]{tcolorbox}
\usepackage{listings}
\newtcolorbox{promptbox}[1][]{
  enhanced, breakable,
  colback=gray!1,
  colframe=gray!60,
  coltitle=black,
  boxrule=2pt,
  arc=10pt,
  left=6pt, right=6pt, top=6pt, bottom=6pt,
  title={#1}, fonttitle=\bfseries,
  attach boxed title to top left={yshift*=-3mm},
  boxed title style={colback=gray!10}
}

\tcbset{
  aibox/.style={
    width=\linewidth,
    top=8pt,
    bottom=4pt,
    colback=reallab-red!15,
    colframe=reallab-red,
    colbacktitle=reallab-red!90!black,
    enhanced,
    center,
    attach boxed title to top left={yshift=-0.1in,xshift=0.15in},
    boxed title style={boxrule=0pt,colframe=white,},
  }
}
\newtcolorbox{AIbox}[2][]{aibox,title=#2,#1}
\usepackage{amsthm}
\usepackage{float}
\theoremstyle{plain}

\theoremstyle{definition}

\theoremstyle{remark}

\definecolor{rqblue}{RGB}{59,111,182}
\definecolor{rqbluebg}{RGB}{234,242,252}
\newtcolorbox{rqbox}[2]{
  enhanced, breakable,
  colback=rqbluebg, colframe=rqblue,
  boxrule=0.9pt, arc=6pt,
  left=8pt, right=8pt, top=7pt, bottom=6pt,
  fonttitle=\sffamily\bfseries\small, coltitle=rqblue,
  title={#1},
  attach boxed title to top left={xshift=10pt, yshift*=-\tcboxedtitleheight/2},
  boxed title style={colback=white, colframe=rqblue, boxrule=0.9pt, arc=5pt},
  before upper={{\itshape\textcolor{rqblue!85!black}{#2}}\par\vspace{3pt}{\color{rqblue!40}\hrule}\vspace{4pt}\textbf{\textcolor{rqblue}{TL;DR.}}\ },
}

\title{
EO-Jev:~Decision Model for Earth Observation 
}

\author[1,\dagger]{Mohamed Atef Abouelhadid}
\author[1,2,\dagger]{Diaa Addeen Abuhani}
\author[1]{Muhammad Akhtar Munir}
\author[1,4]{Fahad Shahbaz Khan}
\author[1,3,*]{Salman Khan}

\affiliation[1]{Mohamed Bin Zayed University of Artificial Intelligence}
\affiliation[2]{Massachusetts Institute of Technology}
\affiliation[3]{Australian National University}
\affiliation[4]{Linköping University Sweden}

\contribution[*]{Corresponding author}
\contribution[\dagger]{Equal Contribution}

\abstract{
Earth observation (EO) workflows require selecting among known outcomes, estimating confidence, and processing large volumes of imagery. Generative vision-language models can address these tasks, but autoregressive decoding is computationally costly and their confidence estimates are unreliable. Decision models offer an alternative, returning probabilities over candidate answers in a single forward pass. However, their accuracy, calibration, and transfer across EO tasks remain insufficiently understood. We investigate these properties through a diagnostic study of d3-flash (8.4B parameters) across 30 settings from 23 public EO benchmarks. We find that compositional reasoning transfers well and calibration improves over its Qwen3.5-9B backbone on 24 settings, while counting, change quantification, damage grading, and absent-object recognition remain challenging. To address these limitations, we adapt d3-flash through low-rank fine-tuning on public EO training splits, producing EO-Jev. EO-Jev improves significantly over d3-flash on 26 of 30 settings and matches or exceeds specialist references on 20 of 28. It also achieves an average gain of 8.0 points across 12 held-out settings and lower expected calibration error than specialists on 21 of 28 settings. These improvements come at a modest cost of 1.8 points in general decision performance. Our results demonstrate the potential of adapting general decision models into competitive EO generalists while highlighting persistent challenges in quantitative perception, option-order consistency, and calibration under distribution shift.
}

\date{Oct 30, 2026}
\metadata[Project Page]{\href{https://mbzuai-oryx.github.io/EO-Jev/}{mbzuai-oryx.github.io/EO-Jev}}
\metadata[GitHub]{\href{https://github.com/mbzuai-oryx/EO-Jev}{mbzuai-oryx/EO-Jev}}
\correspondence{\email{salman.khan@mbzuai.ac.ae}}

\begin{document}

\maketitle
\begin{figure*}[h]
  \centering
  \includegraphics[width=\textwidth]{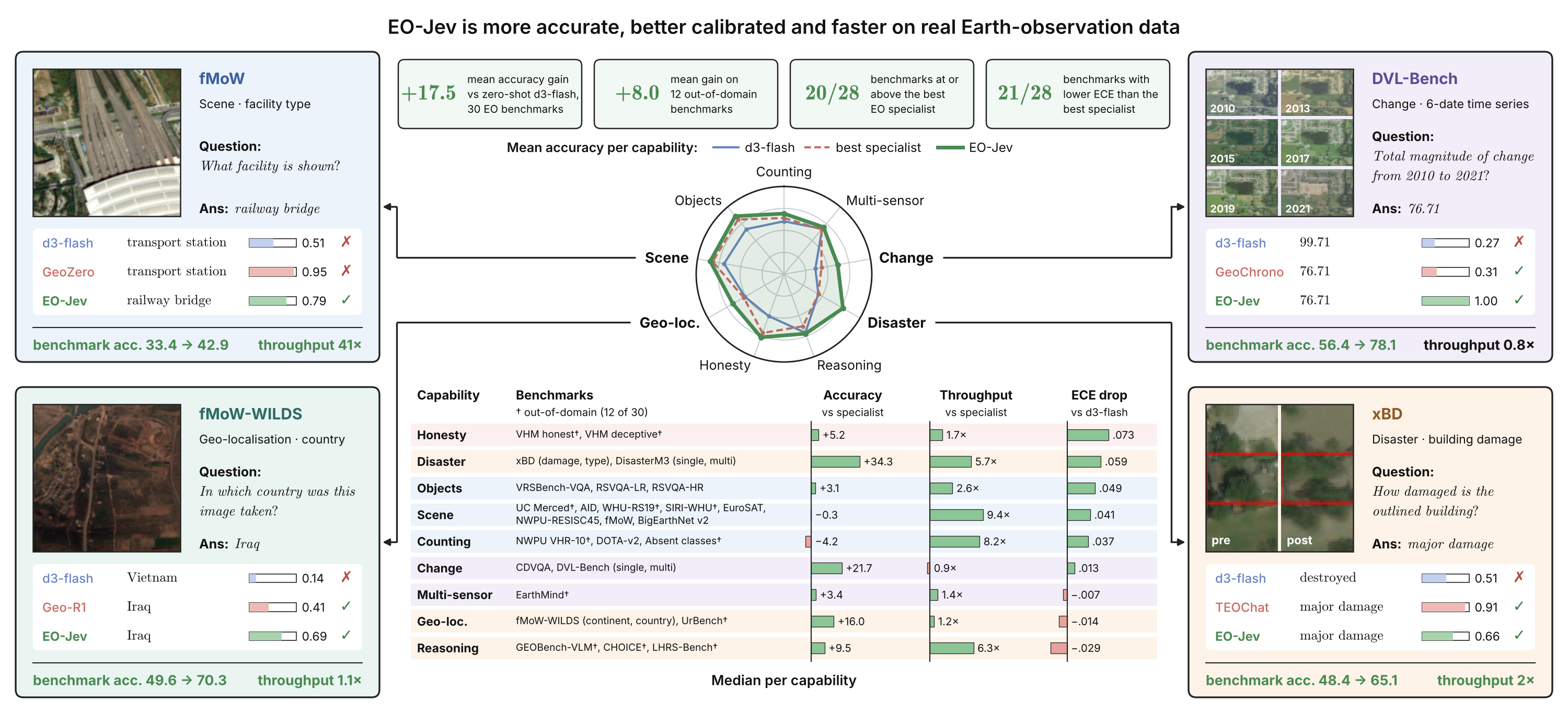}
\caption{\textbf{EO-Jev is more accurate, better calibrated and faster.}
Centre: mean accuracy per capability (30 benchmarks). Corners: real questions with each model's answer
and confidence; footers give benchmark accuracy (specialist $\rightarrow$ EO-Jev) and throughput vs.\ the specialist.
Bottom: per-capability medians of accuracy gain and throughput vs.\ the best specialist, and ECE drop vs.\ d3-flash;
green = better, red = worse.}
  \label{fig:eo-system-one}
\end{figure*}

% =====================================================================
\section{Introduction}
\label{sec:intro}
% =====================================================================

Vision-language models (VLMs) are increasingly used as general-purpose interfaces to EO imagery, answering questions about land use, objects, change and disasters that previously required a dedicated model per task~\citep{geochat,earthdial,vhm,teochat,shabbir2025geopixel}. A new class of models pushes this interface in a different direction. \emph{Decision models}, popularized by Jev~\citep{jev} and its open counterparts~\citep{jev9b,d3flash}, do not generate text. Rather, given a state, a question and a set of options, they return a probability distribution over the options in a single forward pass. For EO applications such as damage triage, monitoring and alerting, this is an attractive capability as an answer that comes with a trustworthy confidence can be thresholded, aggregated and audited, as illustrated in Figure~\ref{fig:eo-system-one}, whereas free-form text generated auto-regressively cannot.

Yet no general decision model has been evaluated on remote-sensing data. The public decision-model leaderboards do not contain EO imagery~\citep{decisionindex}. The closest prior work is RSJEV~\citep{rsjev}, which applies a one-pass decision readout to scene classification using Qwen3.5-0.8B and reports encouraging results on UC Merced, AID, and NWPU-RESISC45. However, it trains and evaluates separately on each benchmark, and its results are not directly comparable with out-of-domain evaluations. Moreover, an independent study shows that letter-readout classifiers can collapse across label vocabularies and confidently misclassify unseen classes~\citep{rsajev}. Therefore, it remains unclear whether a general decision model can perform reliably across EO tasks, how it compares with specialist VLMs, and whether its confidence estimates can be trusted. We address these questions across ten capability groups, with out-of-domain evaluation, EO specialist baselines, and calibration analysis.

In this study, we first conduct a diagnostic study of d3-flash~\citep{d3flash}, an open state-of-the-art 8.4B-parameter decision model derived from Qwen3.5-9B~\citep{qwen35}. We construct 30 closed-form evaluation settings from 23 public benchmarks, covering scene recognition~\citep{ucmerced,aid,nwpu,eurosat}, multi-label land cover~\citep{bigearthnetv2}, visual question answering~\citep{rsvqa,vrsbench}, counting~\citep{dota,nwpuvhr10}, change and temporal reasoning~\citep{cdvqa,dvlbench}, disaster assessment~\citep{xbd,disasterm3}, geo-localization~\citep{fmow,urbench}, complex reasoning~\citep{choice,lhrsbot,geobenchvlm}, and questions about absent objects~\citep{vhm}. The evaluation primarily focuses on optical aerial/satellite imagery. To reduce differences arising from benchmark protocols, we evaluate d3-flash, its Qwen3.5-9B backbone, and selected open-weight EO specialist baselines on identical images, questions, and candidate answers. The backbone uses a comparable answer-code readout, allowing us to examine the effects of decision training under a common evaluation interface.

We then investigate how EO-specific supervision changes the decision model capabilities identified by this analysis. Using low-rank adaptation on public EO training splits, together with general decision-data replay, we obtain \emph{EO-Jev}. The name denotes an EO adaptation of the Jev-style decision interface i.e., a single model that returns probabilities over task-specific candidate answers without autoregressive decoding. We distinguish evaluation settings whose datasets contribute training data during adaptation from settings whose datasets are held out entirely from that stage. This allows us to assess improvements within the adaptation datasets alongside transfer to other datasets. We also examine calibration, option-order sensitivity, inference cost, and retention of general decision capabilities to avoid catastrophic forgetting.

The study addresses four main questions.
\textbf{(RQ1)} Which EO capabilities transfer from a general-purpose decision model without EO-specific adaptation, and how does its performance compare with EO specialists?
\textbf{(RQ2)} How does EO supervision affect performance, and do its benefits extend to datasets held out from adaptation?
\textbf{(RQ3)} How do calibration and inference cost compare with the backbone and specialist baselines?
\textbf{(RQ4)} Which limitations persist after decision model adaptation to EO tasks?

Our results show that d3-flash performs well on several compositional EO reasoning benchmarks and achieves lower expected calibration error than its backbone on 80\% of the cases (24 out of 30 settings).  Its principal weaknesses are counting, change quantification, damage grading, and recognizing absent objects. EO adaptation improves performance significantly on 26 settings (86.7\% cases), with mean gains of $+$23.8 percentage points on settings from datasets used during adaptation and $+$8.0 points on settings from held-out datasets. EO-Jev achieves scores at or above the evaluated specialist reference on 20 of 28 settings and lower expected calibration error on 21 of 28. These improvements are accompanied by a 1.8 point decrease on the general Decision Index, regressions on two reasoning benchmarks, and increased calibration error on some settings. Our overall results characterize the benefits and limitations of EO adaptation while identifying quantitative perception, option-order consistency, and calibration under distribution shift as priorities for further study.

\section{Evaluation Setup}
\label{sec:setup}
% =====================================================================

\paragraph{Models.} d3-flash~\citep{d3flash} is the 8B member of vLLM Semantic Router's Decision~3.0 family. It has 8.39B parameters including a 7.94B text tower, a 0.46B vision tower and a 1.04M-parameter readout. It is built on Qwen3.5-9B, a 32-layer hybrid of Gated DeltaNet and full-attention layers. Instead of a language-modelling head, it applies a $255\times4096$ linear readout to the last prompt token, producing a distribution over up to 255 answer codes. A comparison of its weights with Qwen3.5-9B shows that the vision tower is frozen, the text tower fully fine-tuned, and the full-attention layers made bidirectional over the prompt. As a same-backbone control, we run Qwen3.5-9B through the same runtime with an identical prompt, image processing and answer codes, using its own language-head rows as the readout. Tokenizer, chat template and image processor are byte-identical, so differences between the two isolate decision training. For each benchmark, we select the open-weight EO specialist with the highest reported performance on the corresponding dataset (Table~\ref{tab:specialists}). Each selected specialist is evaluated on identical images, question wording and lettered options. We score both its option probabilities and generated answers, reporting the better result.
\begin{figure*}[!t]
  \centering
  \includegraphics[width=\textwidth]{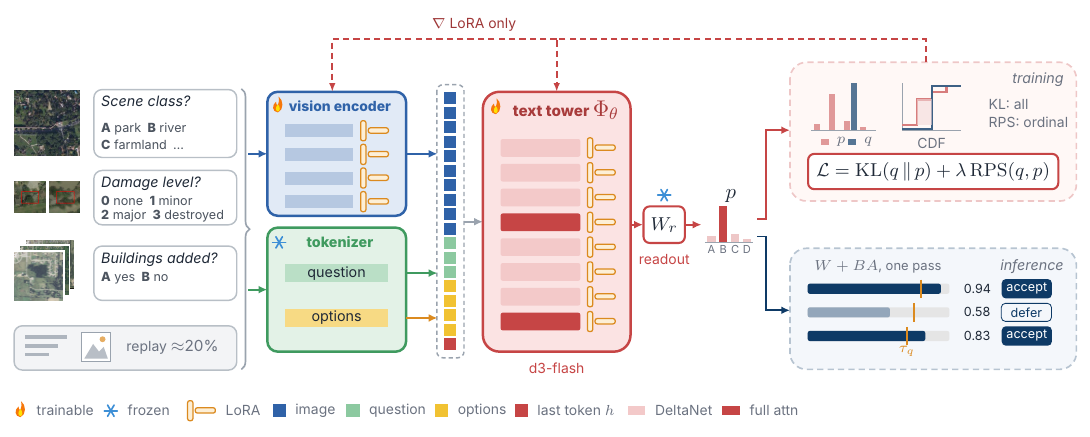}
  \caption{\textbf{EO-Jev Architecture and Training.} EO images (single images, pre/post pairs or time series) are paired with typed questions with predefined answers. General decision data is replayed in 20\% of each batch. The vision encoder and text tower $\Phi_\theta$ process the images and questions, respectively. A frozen readout $W_r$ converts the final hidden state $h$ into option probabilities $p$. \textit{Training:} LoRA adapters in the vision encoder and text tower are optimized using KL divergence against target distributions $q$, with an additional ranked probability score (RPS) on ordinal questions. \textit{Inference:} merged adapters ($W+BA$) enable single-pass predictions at d3-flash cost. Answers are accepted when confidence exceeds a per-question threshold $\tau_q$ and otherwise deferred to an analyst. Examples and probabilities are illustrative.}
  \label{fig:eod3}
\end{figure*}
\paragraph{EO-Jev.} Figure~\ref{fig:eod3} summarizes how we adapt d3-flash to Earth observation. \emph{Training data} (left): EO-Jev is trained on EO decision questions built only from public training splits, covering six families: scene and land-use labels, visual question answering, object counting, multi-temporal change, building damage and disaster type, and geo-localization. Every training image is checked against every evaluation image by perceptual hash, and fMoW training and evaluation images are additionally separated in time and space. To preserve general decision skills, about a fifth of every batch replays general decision data (text-only decision questions and natural-image multiple-choice questions) with d3-flash's own probabilities as targets. \emph{Model} (middle): EO-Jev adds low-rank adapters (LoRA) to the vision blocks (at a lower learning rate), the vision merger and the text tower of d3-flash, and keeps its 255-code readout. The loss matches the target distribution with a KL term, plus a ranked probability score for ordinal questions. Training runs for 20{,}000 steps, after which the adapters are merged into the BF16 weights, so EO-Jev has the same architecture and inference cost as d3-flash. \emph{Decisions} (right): at inference, each typed question returns a probability distribution over its options in a single forward pass, and a per-question threshold decides whether to accept the answer or defer it to an analyst or a specialist model. Section~\ref{sec:details} lists the training sources.

\paragraph{Benchmarks and domains.} The 30 evaluation settings fall into ten capability groups (Table~\ref{tab:results}). Multiple-choice benchmarks are used natively. Yes/no questions become binary decisions, and ordinal targets such as damage levels become ordered-level questions. Multi-select items are scored by exact-set match. We exclude SAR inputs throughout. Each setting is labelled \emph{in-domain} if EO-Jev was trained on other images from the same dataset, and \emph{out-of-domain} if no image from that dataset was used. In total, 18 settings are in-domain and 12 out-of-domain. Every training image was compared with every evaluation image by perceptual hash, so no evaluation image appears in training. For fMoW-WILDS, training and evaluation images are additionally separated in time and space.

\paragraph{Metrics.} We report accuracy, with three exceptions: exact-set match for multi-select settings, strict CircularEval accuracy for LHRS-Bench (correct under all four option orders), and correct abstention for VHM deceptive questions. Calibration is measured by expected calibration error (ECE) with 15 equal-width bins on the top answer's probability. Differences between EO-Jev and d3-flash are tested with a paired 95\% bootstrap over images. As a retention check, we also report the general (non-EO) Decision Index~\citep{decisionindex}.

\begin{table*}[!t]
  \centering
  \caption{\textbf{Earth-observation results.} All models answer identical questions (SAR inputs excluded). The metric is accuracy, except exact-set match on multi-select rows, strict accuracy on LHRS-Bench and correct abstention on VHM deceptive questions. \emph{Best EO specialist}: the strongest open-weight EO model in our common question format. \emph{Domain}: whether EO-Jev was trained on other images from the benchmark's training split. Colours are scaled within each row. $^{\ast}$: the paired 95\% bootstrap interval (over images) of EO-Jev $-$ d3-flash excludes zero. The specialist gap is a plain difference. \emph{Decision Index}: general (non-EO) retention check.}
  \label{tab:results}
  \includegraphics[width=\textwidth]{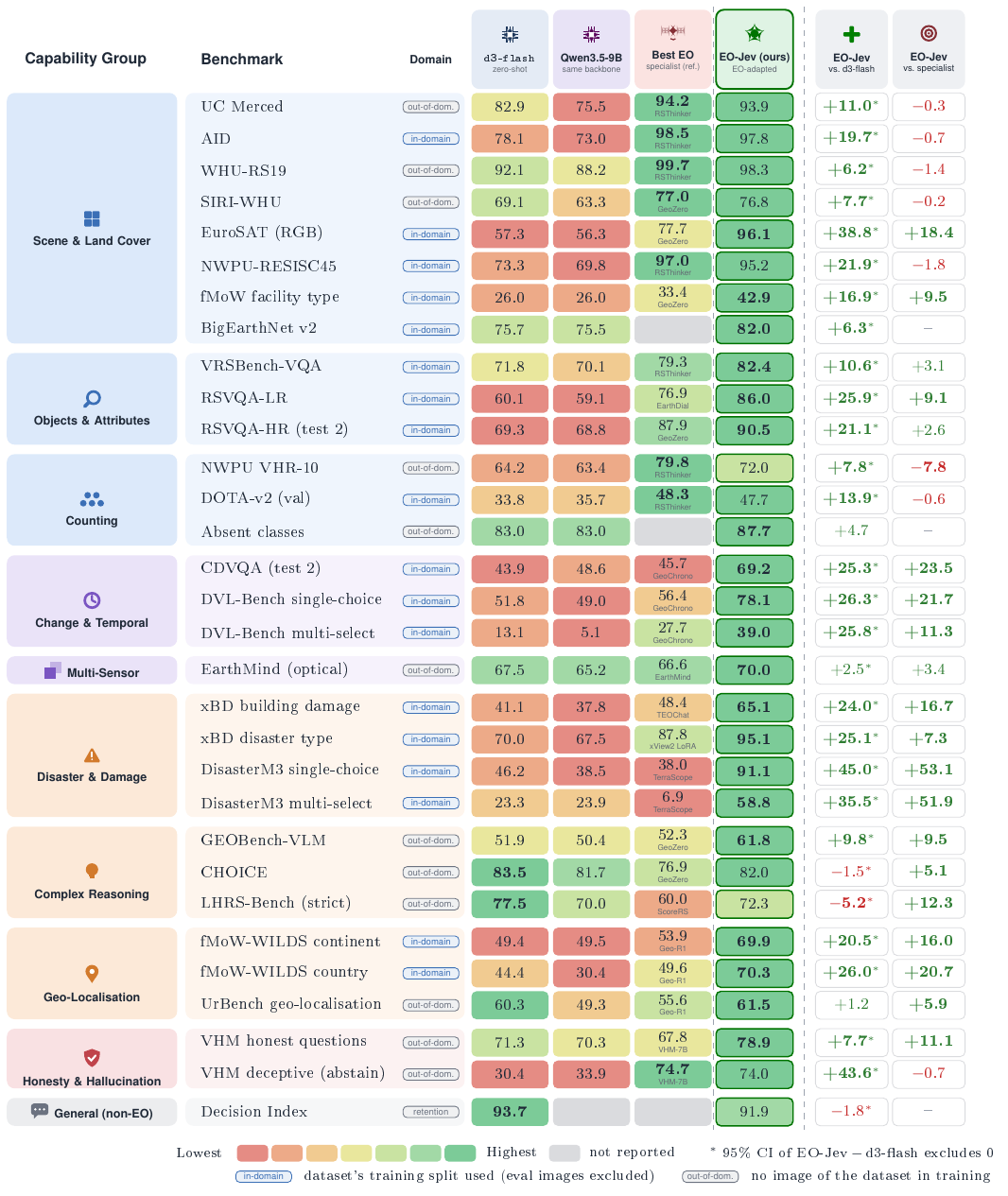}
\end{table*}

% =====================================================================
\section{How Far Does a General Decision Model Reach on EO?}
\label{sec:rq1}
% =====================================================================

\begin{rqbox}{RQ1}{Which EO capabilities transfer from a general-purpose decision model without EO-specific adaptation, and how does its performance compare with EO specialists?}
Without any EO training, d3-flash barely matches or beats the best open EO specialist (7 of 28 settings). These settings test broad reasoning over varied imagery. Specialists lead wherever they were trained on the benchmark, and both d3-flash and its backbone are weak on quantitative perception and abstention.
\end{rqbox}

\paragraph{Compositional reasoning transfers.} The clearest strength of the zero-shot model is broad multiple-choice reasoning over EO imagery. (i)~On LHRS-Bench under strict CircularEval it reaches 77.5, against 60.0 for the best specialist. On CHOICE, whose imagery was collected after most current models were trained, it reaches 83.5 against 76.9. (ii)~It also leads on UrBench geo-localisation (60.3 vs.\ 55.6), on DisasterM3 (46.2 vs.\ 38.0 single-choice; 23.3 vs.\ 6.9 multi-select), on EarthMind optical (67.5 vs.\ 66.6) and on VHM honest questions (71.3 vs.\ 67.8). (iii)~It is roughly tied with GeoZero on GEOBench-VLM (51.9 vs.\ 52.3) and with GeoChrono on CDVQA (43.9 vs.\ 45.7). Most of these benchmarks lie outside the specialists' training data, so they measure transfer rather than fit.

\paragraph{Recognition and visual question answering.} d3-flash trails specialists trained on the corresponding benchmarks, particularly in scene classification and visual question answering. On AID and NWPU-RESISC45, RSThinker achieves 98.5 and 97.0, respectively, compared with 78.1 and 73.3 for d3-flash. On RSVQA, GeoZero and EarthDial outperform d3-flash by 18.6 and 16.8 percentage points. Performance is also weak on low-resolution images and classification tasks with many classes, reaching only 57.3 on $64^2$-pixel EuroSAT tiles and 26.0 on 62-class fMoW facility classification.

\paragraph{Counting, change quantification, damage assessment, and abstention.} These tasks expose several of d3-flash's main limitations. Counting accuracy is 33.8 on DOTA-v2 and 64.2 on NWPU VHR-10. For change detection, the model can identify whether a change has occurred, but struggles to quantify the type of change: on CDVQA, it answers ``buildings'' for 401 of 468 questions asking which class changed most, regardless of the image. On xBD, damage classification accuracy is 41.1, compared with a chance level of 25.0. On DVL-Bench, exact-set accuracy for multi-temporal multi-select questions is just 13.1. Finally, on VHM questions designed to test whether the model recognizes absent objects, it abstains correctly only 30.4\% of the time, below the 33.9\% achieved by its Qwen3.5-9B backbone.

\paragraph{What decision training contributes.} Compared with its Qwen3.5-9B backbone evaluated using an equivalent answer-code readout, d3-flash achieves higher accuracy in 23 of 30 settings, although the gains are generally modest. This suggests that most of its EO perception capabilities are inherited from the backbone. The largest improvements occur in geo-localization, with gains of 14.0 percentage points on fMoW-WILDS country classification and 11.0 on UrBench, and in option-order robustness, with a 7.5-point gain in strict LHRS-Bench accuracy. In earlier zero-shot experiments, decision training also reduced the proportion of GEOBench-VLM answers that changed when the options were reordered, from 40.9\% to 13.5\%. The clearest benefit is improved confidence calibration where d3-flash achieves lower ECE than its backbone in 24 of 30 settings.
(Section~\ref{sec:rq3}).

\paragraph{Main takeaway.} A general decision model already brings two things EO specialists lack. First, compositional reasoning that transfers to new benchmarks, and second, calibrated probabilities. It lacks the quantitative, EO-specific perception that specialists acquire from in-domain training.

% =====================================================================
\section{What Does EO Supervision Add?}
\label{sec:rq2}
% =====================================================================
\begin{figure*}[htbp]
  \centering
  \includegraphics[width=\textwidth]{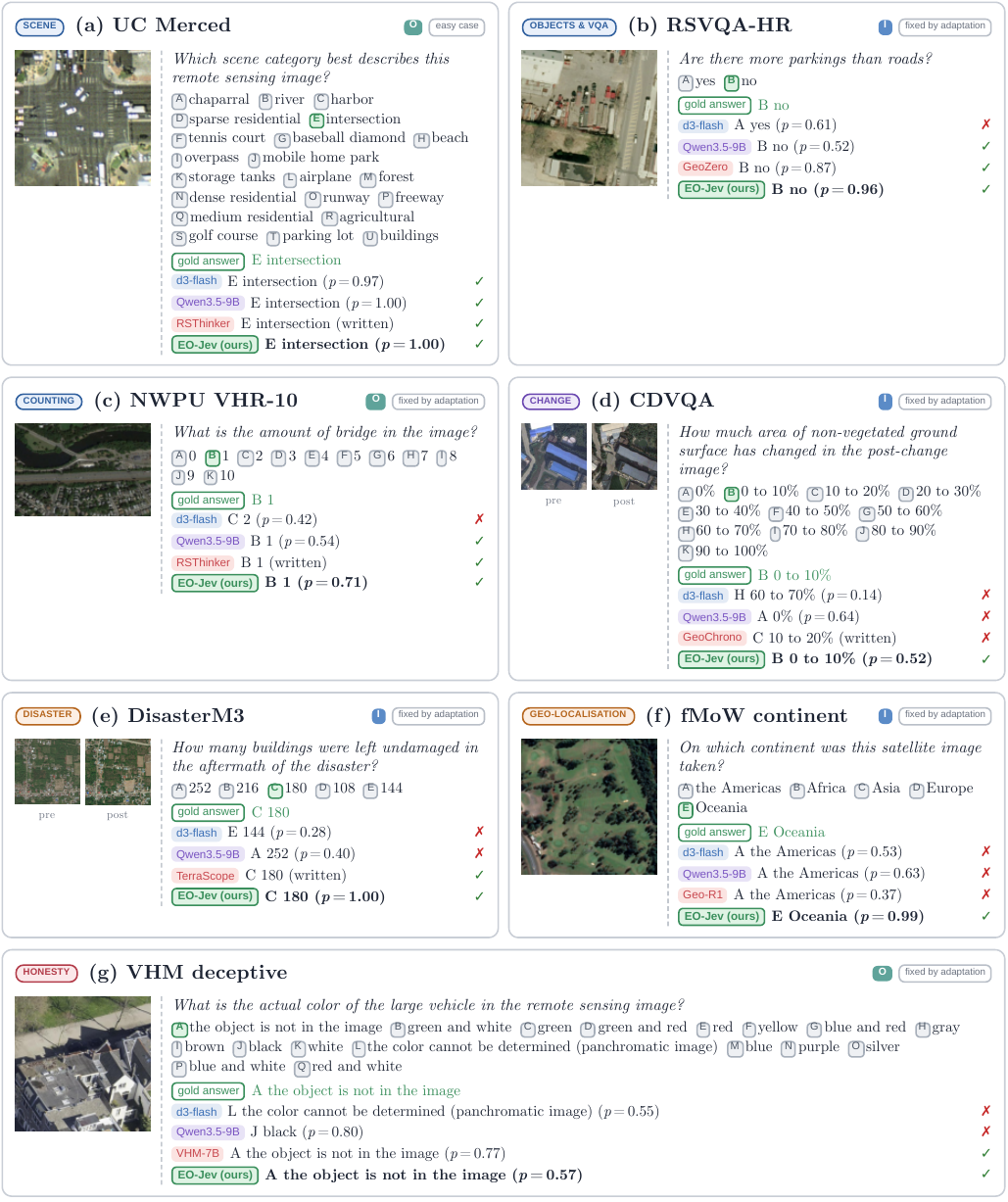}
\caption{\textbf{Qualitative examples across capability groups.} Each panel shows the input image(s), the question
and its options (gold option highlighted in green), and the top answer of each model with its probability, or
``written'' for specialists whose best mode is a generated answer. Badges: I = in-domain, O = out-of-domain for EO-Jev.
In (b)--(g), d3-flash sometimes answers incorrectly with high confidence, defined as assigning a probability of at least 0.8 to the selected answer (e.g., predicting \emph{the Americas} for an Oceanian site in (f)), \emph{the color cannot be determined} in (g)), while EO-Jev selects the correct option; in (d) and (f) the
best specialist is wrong as well. This includes two out-of-domain settings, NWPU VHR-10 counting (c) and VHM deceptive
questions (g), where no training image came from the benchmark's dataset. Panel (a) is an out-of-domain scene that
every model answers correctly, EO-Jev with $p = 1.00$. All images, questions and probabilities are taken unchanged
from the evaluation runs.}
  \label{fig:qa}
\end{figure*}

\begin{rqbox}{RQ2}{How does EO supervision affect performance, and do its benefits extend to datasets held out from adaptation?}
EO supervision improves EO-Jev significantly on 26 of 30 settings: by $+23.8$ points on average in-domain and $+8.0$ out-of-domain. Under equal supervision, the single model matches or beats the specialists on 20 of 28 settings. The costs are small regressions on two option-order-sensitive reasoning benchmarks and 1.8 points of general decision skill. 
\end{rqbox}

\paragraph{In-domain supervision gives large, reliable gains.} All 18 in-domain settings improve significantly, by $+23.8$ points on average (median $+24.6$). Examples include DisasterM3 single-choice ($46.2\rightarrow91.1$), EuroSAT ($57.3\rightarrow96.1$), DisasterM3 multi-select ($23.3\rightarrow58.8$), DVL-Bench, CDVQA, RSVQA-LR and fMoW-WILDS country (each $+25$ to $+26$), xBD damage grading ($41.1\rightarrow65.1$) and xBD disaster type ($70.0\rightarrow95.1$). The fMoW gains show that this is not memorization: training images predate 2013, evaluation images date from 2016 onwards, and training images near any evaluation site are removed, yet country accuracy rises from 44.4 to 70.3. 

\paragraph{The gains transfer beyond the training datasets.} Of the 12 out-of-domain settings, 8 improve significantly, by $+8.0$ points on average across all 12. Transfer is strongest when an unseen benchmark shares a task with the training data. (i)~Scene datasets never used in training improve by 6 to 11 points: UC Merced $+11.0$, SIRI-WHU $+7.7$, WHU-RS19 $+6.2$. (ii)~GEOBench-VLM improves by $+9.8$ and NWPU VHR-10 counting by $+7.8$. (iii)~The largest out-of-domain gain is in abstention: correct abstention on VHM deceptive questions rises from 30.4 to 74.0 even though no VHM data was used, because training on questions about absent objects in other datasets generalizes.

\paragraph{Under equal supervision, one model matches the specialists.} EO-Jev matches or exceeds the best specialist on 20 of 28 settings, and on 14 of the 17 in-domain settings that have one. In most of these comparisons the specialist was also trained on the benchmark's own split, so the comparison is between equally supervised models. Where specialists remain ahead, the margin is under 2 points except on NWPU VHR-10 counting ($-7.8$): UC Merced $-0.3$, AID $-0.7$, WHU-RS19 $-1.4$, SIRI-WHU $-0.2$, NWPU-RESISC45 $-1.8$, DOTA-v2 $-0.6$ and VHM deceptive $-0.7$. The largest leads are on tasks where specialists generalize poorly: DisasterM3 ($+53.1$ and $+51.9$ over TerraScope), CDVQA ($+23.5$) and DVL-Bench ($+21.7$) over GeoChrono, fMoW-WILDS country ($+20.7$) over Geo-R1, EuroSAT ($+18.4$) and xBD damage ($+16.7$).

\paragraph{What adaptation costs.} Four out-of-domain settings do not show a statistically significant improvement.. CHOICE ($-1.5$) and LHRS-Bench strict ($-5.2$) decrease significantly, though both remain well above every specialist. UrBench ($+1.2$) and absent-class questions ($+4.7$) change insignificantly. All four test broad reasoning or robustness rather than one perceptual skill. LHRS-Bench strict accuracy requires a correct answer under every option order, so its decrease points to a slight erosion of the option-order consistency that decision training provides. On the general Decision Index, EO-Jev scores 91.9 against 93.7 for d3-flash, a drop of 1.8 points.

\paragraph{Main takeaway.} EO supervision converts a general decision model into one that rivals specialists across tasks, and its gains reach well beyond the datasets it was trained on. The specialists' edge was supervision rather than architecture: once the general model receives comparable supervision, the edge largely disappears. Figure~\ref{fig:qa} shows one example per capability group, including two out-of-domain cases (counting on NWPU VHR-10 and abstention on VHM deceptive questions) that EO-Jev answers correctly while d3-flash does not.

% =====================================================================
\section{Can the Model's Confidence Be Trusted, and at What Cost?}
\label{sec:rq3}
% =====================================================================

\begin{rqbox}{RQ3}{How do calibration and inference cost compare with the backbone and specialist baselines?}
EO-Jev has lower ECE than d3-flash on 20 of 30 settings, including significant reductions on 17 than its backbone on 27 of 30 and than the best specialist on 21 of 28. Where calibration worsens, the cause is systematic over-confidence. Answering in one forward pass, EO-Jev is a median 6.4$\times$ faster than the best specialist.
\end{rqbox}

\begin{figure}[!t]
  \centering
  \includegraphics[width=\linewidth]{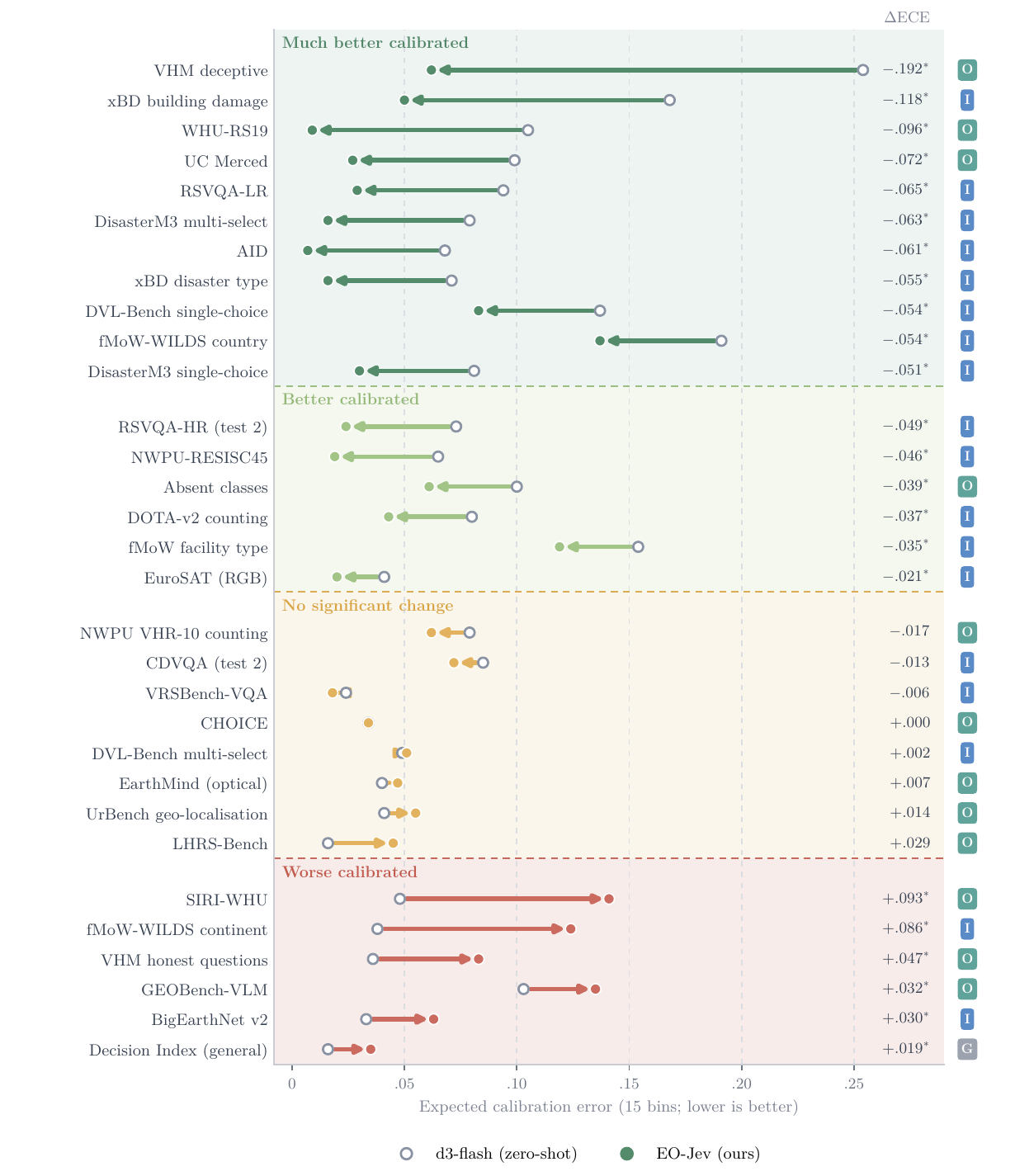}
  \caption{\textbf{Calibration change from d3-flash to EO-Jev.} Each arrow runs from d3-flash's ECE (open circle) to EO-Jev's (filled circle); lower is better. Rows are grouped by $\Delta$ECE $=$ EO-Jev $-$ d3-flash: much better calibrated ($\Delta$ECE $\leq -0.05$), better calibrated, no significant change, and worse calibrated. $^{\ast}$: the paired 95\% bootstrap interval (over images) excludes zero. Badges: \textsf{I} = in-domain, \textsf{O} = out-of-domain, \textsf{G} = general (non-EO) retention check. All EO rows in the bottom tier are over-confident; the Decision Index is slightly under-confident.}
  \label{fig:calibration}
\end{figure}

\paragraph{Calibration improves where accuracy improves most.} ECE is measured over the same questions for every model; multi-select options count as binary questions. As Figure~\ref{fig:calibration} shows, EO-Jev has significantly lower ECE than d3-flash on 17 of 30 EO settings, higher on 5, and no significant change on 8. The largest improvements are on settings where d3-flash was both inaccurate and over-confident. ECE on VHM deceptive questions falls from 0.254 to 0.062, xBD damage grading from 0.168 to 0.050, and WHU-RS19 from 0.105 to 0.009. On in-domain scene and VQA settings, EO-Jev's ECE is typically 0.01--0.03.

\paragraph{Specialists' probabilities are often uninformative.} We evaluate specialists using their option-probability outputs because their generated answers do not provide confidence estimates. EO-Jev achieves lower ECE than the best specialist in 21 of 28 settings and lower ECE than Qwen3.5-9B in 27 of 30. Specialist calibration is competitive on scene classification, where RSThinker achieves ECE values between 0.004 and 0.052. However, calibration is much worse on counting and multi-select tasks. RSThinker reaches an ECE of 0.712 on NWPU VHR-10 and 0.356 on DOTA-v2, while TerraScope reaches 0.529 on DisasterM3 multi-select. In these settings, specialist confidence is a poor indicator of whether an answer is correct.

\paragraph{Calibration errors are systematically overconfident.} Whenever EO-Jev is significantly less calibrated than d3-flash on an EO setting, its mean confidence exceeds its accuracy. This gap reaches 12--14 percentage points on SIRI-WHU, GEOBench-VLM, and fMoW-WILDS continent. It is also 7.3 points on VHM honest questions and 6.2 points on BigEarthNet. The latter two results are associated with a tendency to answer ``yes'' to questions about object presence. Since these errors reflect systematic overconfidence rather than random variation, per-task temperature scaling fitted on a development set may correct much of the miscalibration. We leave this evaluation for future work.

\begin{figure}[t]
  \centering
  \begin{minipage}[t]{0.49\linewidth}\centering
    \includegraphics[width=\linewidth]{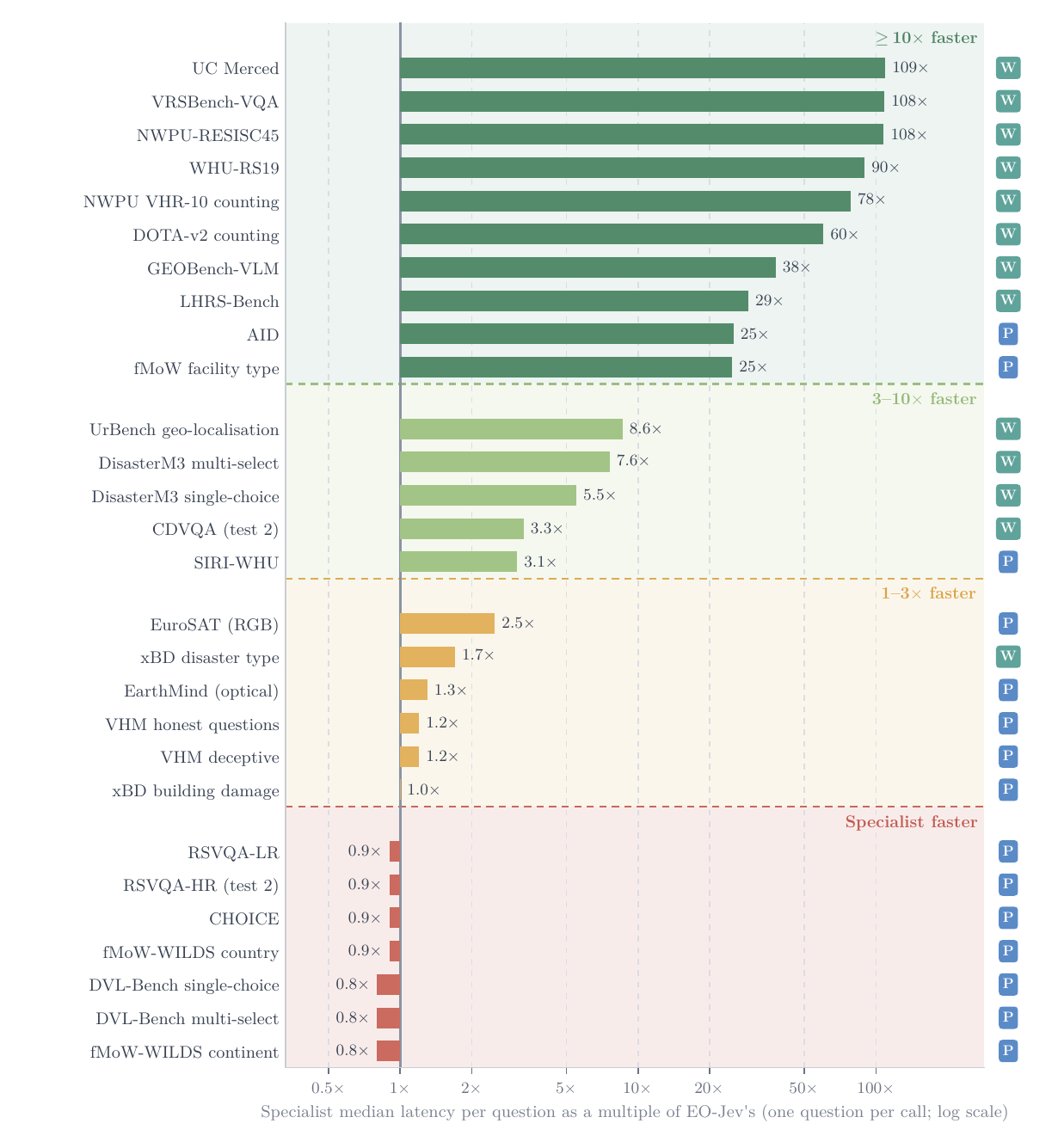}\\[-1pt]
    {\small (a) One question per call}
  \end{minipage}\hfill
  \begin{minipage}[t]{0.49\linewidth}\centering
    \includegraphics[width=\linewidth]{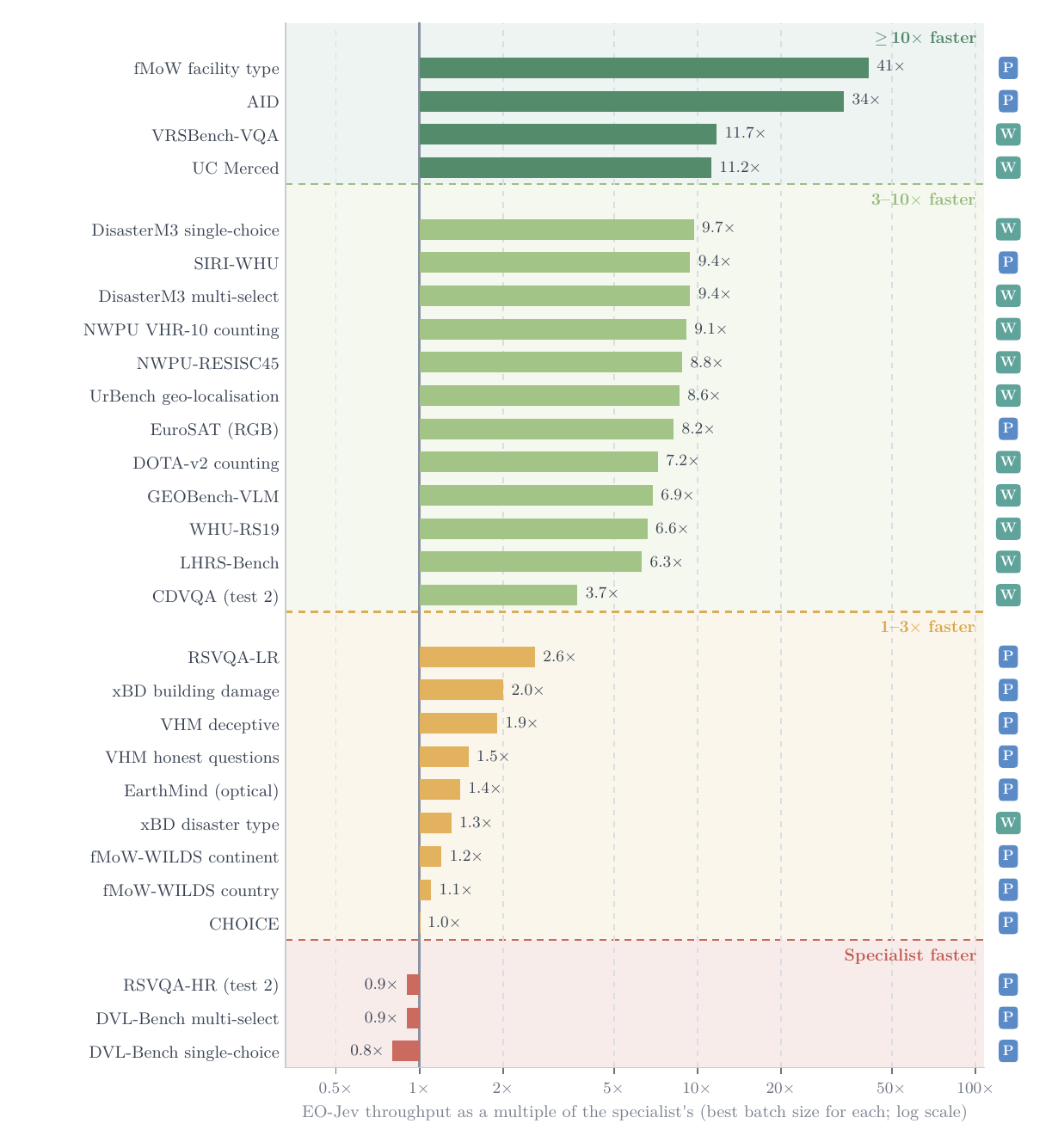}\\[-1pt]
    {\small (b) Best batch size for each model}
  \end{minipage}
  \caption{\textbf{Inference cost against the specialists.}
  (a)~Specialist median latency per question as a multiple of EO-Jev's, one question per call.
  (b)~EO-Jev throughput as a multiple of the specialist's, each at its best batch size.
  Bars are grouped by speed-up; red rows are those where the specialist is faster.
  Badges give the specialist's best-accuracy mode (P: prompted, W: wrapped/generative).
  All models run on the same exclusive node (4$\times$A100, BF16), on the same 160-question seeded sample per row, with image decoding inside the timed region.
  Medians: 3.2$\times$ in (a) and 6.4$\times$ in (b).}
  \label{fig:time}
\end{figure}

\paragraph{One forward pass is much cheaper than answer generation.} With one question per call, EO-Jev is a median 3.2$\times$ faster than the best specialist (range 0.8--109$\times$); the specialist is faster on 7 of 28 rows and ties on one. With batching the median rises to 6.4$\times$ (range 0.8--41$\times$), and the specialist is faster on only 3 rows (See Figure \ref{fig:time}. The large gains come almost entirely from generative specialists that must decode an answer. Against prompted VLMs of similar size (RSVQA, DVL-Bench, fMoW-WILDS, CHOICE), EO-Jev costs about the same, since both run one forward pass. The speed advantage is therefore structural: it comes from reading out a decision instead of generating one, not from a smaller model.

\paragraph{Main takeaway.} Decision models provide a practical combination of calibrated confidence and low inference cost for automated EO pipelines. They produce answers in a single forward pass, while specialists that generate written answers are several times slower. Specialists evaluated through option probabilities can match this speed, but their confidence is often poorly calibrated, particularly on counting and multi-select tasks.

% =====================================================================
\section{What Remains Hard?}
\label{sec:rq4}
% =====================================================================

\begin{rqbox}{RQ4}{Which limitations persist after decision model adaptation to EO tasks?}
The remaining gaps are counting on unseen imagery, robustness to option order, geometric and cross-view reasoning, and calibration far from the training data. Zero-shot diagnostics show that recognition failures were mostly a readout problem, which adaptation fixed, while quantitative failures reflect missing perception.
\end{rqbox}

\paragraph{Counting on unseen imagery.} Counting is the only capability where a specialist still leads clearly. On out-of-domain NWPU VHR-10, EO-Jev improves from 64.2 to 72.0 but trails RSThinker by 7.8 points. On in-domain DOTA-v2 it reaches 47.7, within 0.6 points of RSThinker, but absolute accuracy remains low for every model. Zero-shot, extra resolution helped counting slightly: tiling ultra-high-resolution scenes improved counting by 5.0 points. Counting therefore appears limited by perception at the 1.6\,MP input budget as well as by supervision.

\paragraph{Robustness to option order.} d3-flash's decision training made it far more stable under option reordering than its backbone. Adaptation slightly erodes this: CHOICE drops by 1.5 points and LHRS-Bench strict accuracy by 5.2. Training with option-permutation augmentation, and keeping option-order tests as regression checks, is the obvious remedy.

\paragraph{Geometric and cross-view reasoning.} UrBench does not change significantly with adaptation ($60.3\rightarrow61.5$). Zero-shot, its orientation-estimation questions were near chance (30.5 against a 24\% random baseline). Our training data contains no cross-view supervision. Zero-shot diagnostics also show orientation sensitivity on single images: about a fifth of correct scene and GEOBench-VLM answers change under some rotation, even though nadir imagery has no canonical up direction.

\paragraph{Calibration away from the training distribution.} The settings where EO-Jev becomes over-confident (SIRI-WHU, GEOBench-VLM, fMoW-WILDS continent, VHM honest questions) are mostly out-of-domain, or involve label sets with many similar classes. Adaptation sharpens the output distributions everywhere, but accuracy improves less out of domain.

\paragraph{Readout versus representation.} Zero-shot probes explain why recognition responds so well to adaptation. A linear probe on d3-flash's decision state, the hidden state its readout scores, reaches 96.4, 94.3 and 87.5 on AID, EuroSAT and NWPU-RESISC45, where its zero-shot answers reach 78.1, 56.8 and 73.5. The information needed to recognise the scene was already present but not mapped to the right label. Adaptation recovers almost exactly this gap: EO-Jev reaches 97.8, 96.1 and 95.2 on the evaluation sets of the same datasets. Counting and change quantification were weak in both the zero-shot model and its backbone, and they improve less. This suggests that these skills need new perceptual supervision, not only a better readout.

% =====================================================================
\section{Where Does a Decision Model Stand on EO?}
\label{sec:standing}
% =====================================================================

\begin{figure}[!t]
  \centering
  \includegraphics[width=\linewidth]{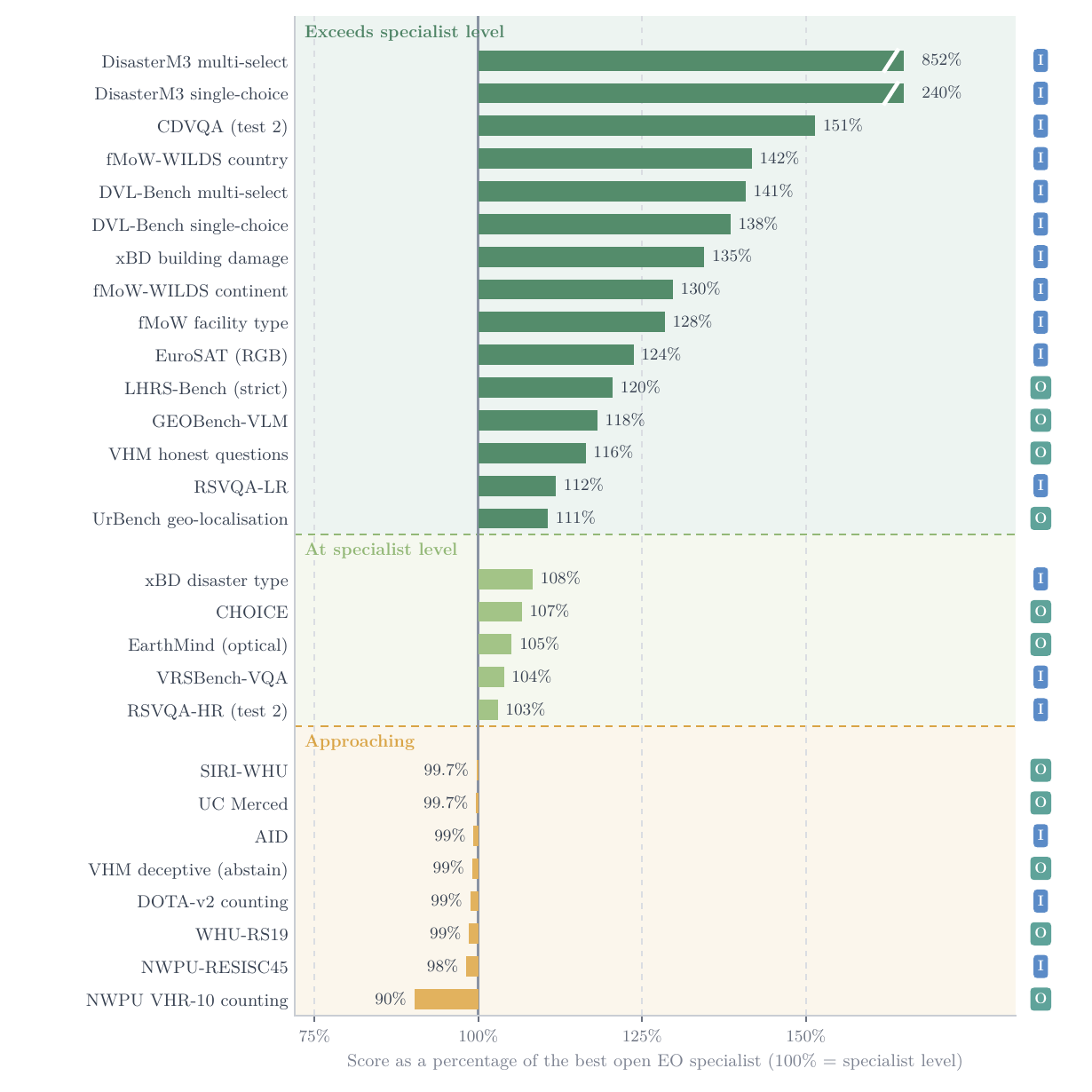}
  \caption{\textbf{EO capability relative to specialist level.} For each setting with a specialist reference, EO-Jev's score as a percentage of the best open EO specialist's. Settings are grouped into tiers: exceeds specialist level (above 110\%), at specialist level (100--110\%) and approaching (85--100\%). No setting falls below 85\%. Bars beyond 165\% are clipped. Badges: \textsf{I} = in-domain, \textsf{O} = out-of-domain.}
  \label{fig:maturity}
\end{figure}

Figure~\ref{fig:maturity} summarizes where EO-Jev stands relative to the strongest available open EO specialist on each setting. Fifteen settings exceed specialist level by more than 10\%. They are dominated by tasks where specialists generalize poorly or were trained for a different question format: disaster assessment, change and temporal reasoning, geo-localization and fine-grained recognition. Five more are at specialist level. The remaining eight are approaching it, all within 2\% except NWPU VHR-10 counting (90\%). They comprise five scene datasets, two counting settings, and deceptive-question abstention, where VHM-7B was trained for exactly that task. No setting shows a substantial gap.

These results suggest a practical division of labor. A single calibrated decision model can serve as the default interface for EO decisions: one forward pass per question, confidence that supports thresholding and deferral, and accuracy at or above specialist level on most tasks. Specialist models remain valuable where a task needs perception the decision model has not learned, such as counting on unfamiliar imagery, and where their generated explanations matter more than calibrated confidence. Because a decision model's errors come with confidence, a pipeline can route its uncertain cases to specialists or analysts rather than replace either.

% =====================================================================
\section{Additional Evaluation Details}
\label{sec:details}
% =====================================================================

\paragraph{Benchmarks.} Scene classification: UC Merced~\citep{ucmerced}, AID~\citep{aid}, WHU-RS19~\citep{whurs19}, SIRI-WHU~\citep{siriwhu}, EuroSAT (RGB)~\citep{eurosat}, NWPU-RESISC45~\citep{nwpu} and fMoW facility type~\citep{fmow}. Multi-label land cover: BigEarthNet v2~\citep{bigearthnetv2}. Object presence and attributes: VRSBench~\citep{vrsbench}, RSVQA-LR and RSVQA-HR~\citep{rsvqa}. Counting: NWPU VHR-10~\citep{nwpuvhr10} and DOTA-v2 validation~\citep{dota}, plus absent-class questions on NWPU VHR-10 images. Change and multi-temporal: CDVQA~\citep{cdvqa} and DVL-Bench~\citep{dvlbench}. Multi-sensor: EarthMind-Bench, optical part only~\citep{earthmind}. Disaster and damage: xBD~\citep{xbd} and DisasterM3~\citep{disasterm3}. Complex reasoning: GEOBench-VLM~\citep{geobenchvlm}, CHOICE~\citep{choice} and LHRS-Bench~\citep{lhrsbot}. Geo-localization: fMoW-WILDS~\citep{fmow,wilds} and UrBench~\citep{urbench}. Honesty: VHM-HnstD~\citep{vhm}.

\paragraph{Output processing.} d3-flash, Qwen3.5-9B and EO-Jev return a probability for every option; the prediction is the most probable option. Every specialist sees the same question text, built from our converted rows; only its chat template differs. In option-probability mode, a specialist's answer is pre-filled up to where it writes the option, and each option's probability is read from the next-token distribution. In generation mode, the specialist decodes greedily in its own reasoning format and the answer is mapped to an option. Before trusting a specialist's scores, we reproduced one of its published numbers where our protocol matches the original.

\begin{table}[!t]
\centering
\small
\caption{\textbf{Specialist reference models.} The strongest open-weight EO specialist per setting in our common question format.}
\label{tab:specialists}
\begin{tabularx}{\linewidth}{@{}l>{\raggedright\arraybackslash}X l@{}}
\toprule
Capability & Settings & Specialist \\
\midrule
Scene classification & UC Merced, AID, WHU-RS19, NWPU-RESISC45 & RSThinker~\citep{rsthinker} \\
 & SIRI-WHU, EuroSAT, fMoW facility type & GeoZero~\citep{geozero} \\
Objects \& attributes & VRSBench & RSThinker~\citep{rsthinker} \\
 & RSVQA-LR & EarthDial~\citep{earthdial} \\
 & RSVQA-HR & GeoZero~\citep{geozero} \\
Counting & NWPU VHR-10, DOTA-v2 & RSThinker~\citep{rsthinker} \\
Change \& temporal & CDVQA, DVL-Bench & GeoChrono~\citep{geochrono} \\
Multi-sensor & EarthMind (optical) & EarthMind~\citep{earthmind} \\
Disaster \& damage & xBD building damage & TEOChat~\citep{teochat} \\
 & xBD disaster type & xView2 LoRA~\citep{waybob2025qwen3vl_xview2} \\
 & DisasterM3 & TerraScope~\citep{terrascope} \\
Complex reasoning & GEOBench-VLM, CHOICE & GeoZero~\citep{geozero} \\
 & LHRS-Bench & ScoreRS~\citep{scorers} \\
Geo-localisation & fMoW-WILDS, UrBench & Geo-R1~\citep{geor1} \\
Honesty & VHM-HnstD & VHM-7B~\citep{vhm} \\
\bottomrule
\end{tabularx}
\end{table}

\paragraph{Training data for EO-Jev.} Training examples are drawn exclusively from public training splits, with all evaluation images removed to prevent data leakage. For scene classification and land-use mapping, we use AID images outside the evaluation list, the EuroSAT and BigEarthNet v2 training splits, and fMoW-WILDS training images acquired before 2013 and located away from evaluation sites. Object and building counting examples come from DOTA-v2 training tiles, excluding the validation split, and building counts derived from xBD pre-disaster footprints. Damage assessment and disaster-type classification examples are drawn from the xBD train and tier3 splits. For general EO reasoning, we use the training splits of NWPU-RESISC45, RSVQA, VRSBench, CDVQA, DisasterM3, and DVL-Bench. Finally, general decision-making is supported by text-only decision questions and natural-image multiple-choice questions, using d3-flash's own probabilities as training targets.

% \paragraph{Timing.} Times are GPU seconds per 1{,}000 questions, from each record's wall-clock time. d3-flash, Qwen3.5-9B and EO-Jev time one image set per call with all its questions batched, after image decoding. Specialists are timed over groups of rows with image decoding included, which slightly favors EO-Jev. d3-flash and EO-Jev were timed on different clusters, so we report a single cost for both: their computation is identical.

\paragraph{Timing protocol.}
We time every system on one exclusive node (4$\times$A100-80GB, pinned CPU cores, no co-scheduled jobs), in BF16, after a warm-up pass. Each row uses the same seeded sample of 160 questions.
The clock covers image loading and decoding, preprocessing, the forward pass and, for generative specialists, answer decoding and parsing; we synchronize CUDA before reading it. We report two regimes; median latency with one question per call, and peak throughput with each model at its own best batch size. EO-Jev and d3-flash are timed separately, and their median latencies are identical (ratio 1.00) given that the merged LoRA adds no cost at inference.

% \paragraph{Relation to RSJEV.} RSJEV~\citep{rsjev} applies the same one-pass decision readout to remote-sensing scene classification with Qwen3.5-0.8B, training and testing within each of UC Merced, AID and NWPU-RESISC45. It reports 97.71, 96.56 and 94.59 accuracy, against 93.9 (out-of-domain), 97.8 and 95.2 for EO-Jev. These numbers are not directly comparable: the test splits and amount of training data differ, RSJEV's AID and NWPU-RESISC45 test images include images EO-Jev was trained on, and UC Merced is in-domain for RSJEV but out-of-domain for EO-Jev. RSJEV independently supports our premise that a decision readout outperforms generative answering on the same backbone. Our study extends it to ten capability groups, out-of-domain evaluation, EO specialist baselines, calibration on every setting and a forgetting check.

% =====================================================================
\section{Conclusion}
\label{sec:conclusion}
% =====================================================================

We studied decision models for earth observation through a diagnostic evaluation across 30 settings from 23 public benchmarks, followed by EO-specific adaptation. The evaluation shows that d3-flash transfers well to several compositional reasoning tasks and generally improves calibration over its backbone, but remains limited in counting, change quantification, damage grading, and recognizing absent objects. Our proposed adaptation approach yields EO-Jev, which improves significantly on 26 settings and achieves scores at or above the evaluated specialist reference on 20 of 28. Its average gain of 8.0 percentage points across 12 settings held out from adaptation demonstrates that the benefits extend beyond the training datasets. EO-Jev also achieves lower expected calibration error than specialist references on 21 of 28 settings while retaining single-pass prediction without autoregressive decoding.

These gains come with trade-offs. Adaptation reduces general decision performance by a small margin of 1.8 points, lowers accuracy on two reasoning benchmarks, and increases calibration error on some settings. Counting on unfamiliar imagery and geometric reasoning also remain difficult. The findings support decision models as an efficient, shared interface for diverse EO tasks, while showing that accuracy, calibration, and robustness must be assessed separately. Further progress requires stronger quantitative and geometric perception, consistency under option reordering, and evaluation of confidence-based deferral under distribution shift. Beyond standalone prediction, integrating calibrated decision models into agentic EO pipelines offers a promising direction for building systems that can reason across tasks while identifying when their predictions require further verification.

\bibliographystyle{plainnat}
\bibliography{references}

\end{document}